\pdfoutput=1

\documentclass[11pt]{article}

\usepackage[final]{acl}

\usepackage{times}
\usepackage{latexsym}

\usepackage[T1]{fontenc}

\usepackage[utf8]{inputenc}

\usepackage{microtype}

\usepackage{inconsolata}

\usepackage{graphicx}

\usepackage{amsmath}

\DeclareMathOperator*{\argmin}{arg\,min}
\usepackage{multirow}
\usepackage{multicol}
\usepackage{makecell}
\usepackage{pifont}

\title{Exploring In-Image Machine Translation with Real-World Background}

\author{
    Yanzhi Tian\textsuperscript{1} ~~~ Zeming Liu\textsuperscript{2} ~~~ Zhengyang Liu\textsuperscript{1} ~~~ Yuhang Guo\textsuperscript{1}\thanks{Corresponding author}
    \\
    \textsuperscript{1}School of Computer Science and Technology, Beijing Institute of Technology
    \\ 
    \textsuperscript{2}School of Computer Science and Engineering, Beihang University
    \\
    \texttt{\normalsize \href{mailto:tianyanzhi@bit.edu.cn}{tianyanzhi@bit.edu.cn}} ~~~ \texttt{\normalsize \href{mailto:zmliu@buaa.edu.cn}{zmliu@buaa.edu.cn}}
    \\
    \texttt{\normalsize \href{mailto:zhengyang@bit.edu.cn}{zhengyang@bit.edu.cn}} ~~~ \texttt{\normalsize \href{mailto:guoyuhang@bit.edu.cn}{guoyuhang@bit.edu.cn}}
}

\begin{document}
\maketitle
\begin{abstract}
In-Image Machine Translation (IIMT) aims to translate texts within images from one language to another. Previous research on IIMT was primarily conducted on simplified scenarios such as images of one-line text with black font in white backgrounds, which is far from reality and impractical for applications in the real world.
To make IIMT research practically valuable, it is essential to consider a complex scenario where the text backgrounds are derived from real-world images.
To facilitate research of complex scenario IIMT, we design an IIMT dataset that includes subtitle text with real-world background.
However previous IIMT models perform inadequately in complex scenarios.
To address the issue, we propose the DebackX model, which separates the background and text-image from the source image, performs translation on text-image directly, and fuses the translated text-image with the background, to generate the target image. 
Experimental results show that our model achieves improvements in both translation quality and visual effect.\footnote{Code and dataset: \url{https://github.com/BITHLP/DebackX}}
\end{abstract}

\section{Introduction}
In-Image Machine Translation (IIMT) aims to translate texts within images from one language to another, and related technologies are widely used in translation applications \cite{mansimov-etal-2020-towards, tian-etal-2023-image, lan2024translatotronv}.
IIMT performs translation on visual modality, helping people understand context in images directly, which is different from the text-to-text machine translation.

\begin{figure}[t!]
    \centering
    \includegraphics[width=\linewidth]{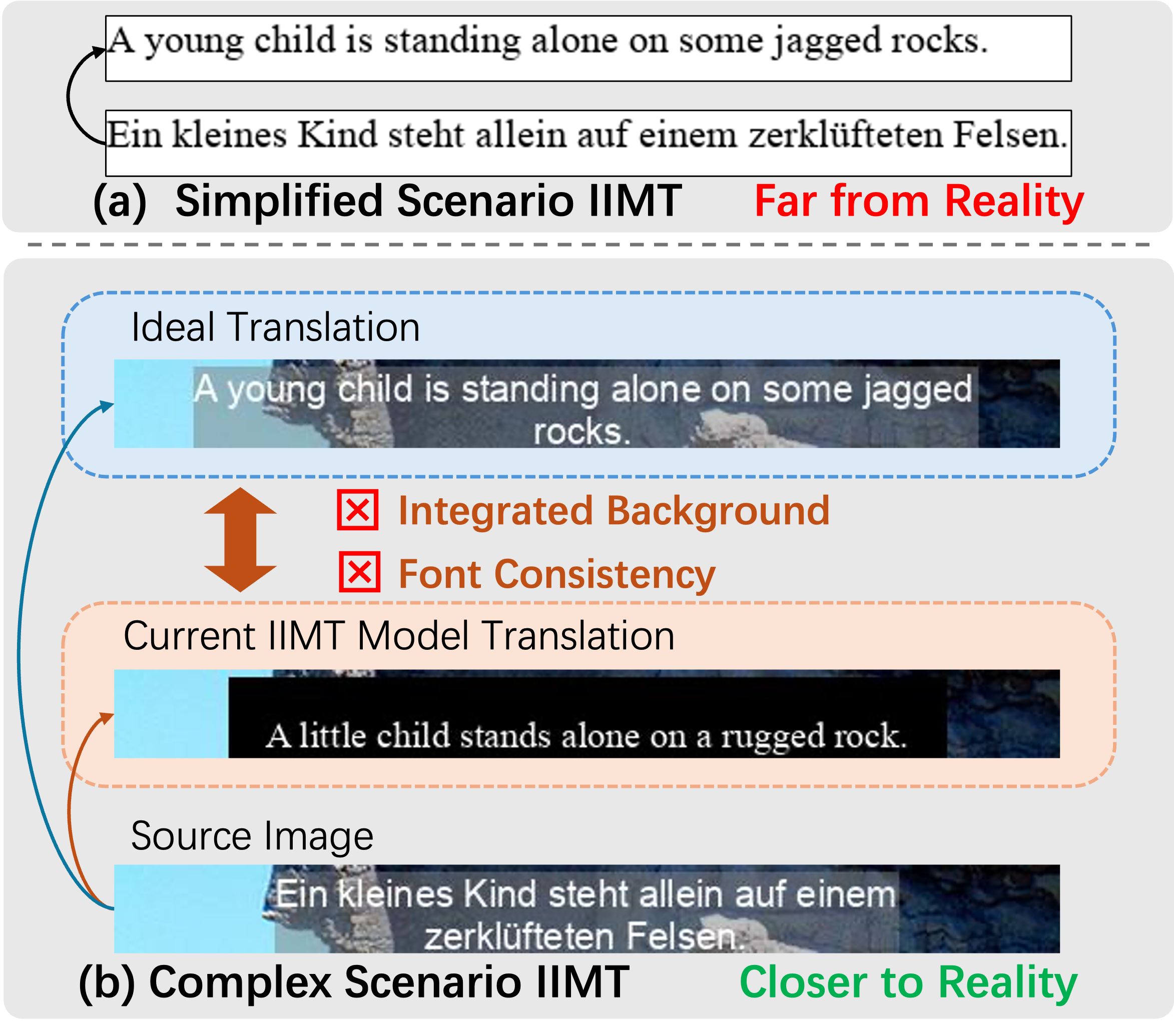}
    \caption{Illustration of simplified and complex scenario IIMT. Previous research mainly focuses on simplified scenario IIMT, which is far from reality. Exploring complex IIMT scenarios that are much closer to reality is necessary. We find that the current IIMT model cannot fully handle the complex scenario, since the translation lacks an integrated background and fails to maintain font consistency, leading to poor visual effect.}
    \label{fig:example1}
\end{figure}

IIMT is a challenging machine translation task, and previous research mainly focuses on simplified scenarios as shown in Figure \ref{fig:example1}(a).
\citet{tian-etal-2023-image} translate images containing texts with black font and white background. 
\citet{lan2024translatotronv} construct images with a single color background.
However, the aforementioned research overly simplifies the IIMT task, ignoring the fact that most of the text in the real world appears with a complex background, such as video subtitles, as shown in Figure \ref{fig:example1}(b). 
The limitation makes previous approaches impractical for applications in the real world, highlighting the need for a scenario that is closer to reality.

To overcome the limitation and make IIMT research practical, we focus on a more realistic IIMT task, namely complex scenario IIMT, building a new IIMT dataset where the images contain text integrated with real-world background.
Compared to the simplified IIMT task that hardly aligns with reality, our designed task better reflects real world.

On the complex scenario dataset, we experiment with a feasible OCR-NMT-Render model, which primarily consists of four steps: 
(1) Recognizing the source language texts with the Optical Character Recognition (OCR) model; (2) Translating the source texts recognized by the OCR model with the Neural Machine Translation (NMT) model; (3) Removing the text areas in the source image to erase the texts; (4) Rendering the translation results into the corresponding position of the image.

The above framework has the following drawbacks. Firstly, the cascading OCR and NMT models have the risk of error propagation, leading to the decreasing translation quality, which is demonstrated in the previous works \cite{tian-etal-2023-image, lan2024translatotronv}.
Secondly, removing text areas in step (3) affects the integrity of the source images, decreasing the visual effect of rendering results.
Thirdly, OCR models generally only give recognized text results, leading to the loss of font information during the cascading process, and it is hard to maintain consistency for font styles of texts between the source images and target images.
The comparison between the ideal translation result and the translation produced by the OCR-NMT-Render model is shown in Figure \ref{fig:example1}(b).

To address the drawbacks, we design the DebackX model for complex scenario IIMT, consisting of three components: 
(i) A Text-Image Background Separation model is used to generate the background and text-image from the source image. Processing the background and text-image separately, our model ensures the integrity of the background.
(ii) An Image Translation model is used to model the transformation from the text-image of the source language to the text-image of the target language. The direct translation mitigates the decreasing translation quality caused by error propagation between OCR and NMT and helps maintain font consistency between the target images and source images. 
(iii) A Text-Image Background Fusion model is used to fuse the background and target text-image to generate the final output.

The main contributions of this paper are as follows:
\begin{itemize}
    \item We propose a complex scenario IIMT, which requires translating text within real-world backgrounds to align closely with reality.
    \item We build a new dataset IIMT30k containing backgrounds derived from real-world images for complex scenario IIMT. Existing IIMT models, including GPT-4o, fail to achieve both high-quality translation and visual effects.
    \item We propose the DebackX model for complex scenario IIMT, processing the background and text of the image separately, achieving both better translation quality and visual effect compared to current IIMT models.
\end{itemize}

\section{Related Work}

\paragraph{Text-Image Translation.}
Conventional research on Neural Machine Translation usually focuses on text and speech modality \cite{vaswani2023attention, inaguma-etal-2023-unity}. 
There is also research focusing on translating the text in the image, which is an image-to-text multi-modality translation task, referred to Text-Image Translation (TIT).
\citet{ma2022improving} construct a synthetic dataset and employ multi-task learning in the training of end-to-end TIT model.
\citet{lan2023exploring} annotate the OCRMT30K dataset, and propose an OCR-NMT cascade TIT model, introducing visual information with a multimodal codebook into the NMT model.
\citet{ma-etal-2023-ccim} propose a cross-model cross-lingual interactive model for TIT with weighted interactive attention and hierarchical interactive attention.
\citet{zhu-etal-2023-peit} design a shared encoder-decoder backbone with vision-text representation aligner and cross-model regularizer.

\paragraph{In-Image Machine Translation.}
All of the above works only generate the translated texts but do not generate the images with translation.
\citet{mansimov-etal-2020-towards} explore the In-Image Machine Translation (IIMT) task, aiming to transform images containing texts from one language to another, which is closer to the real-world applications. 
Both the input and output of IIMT are entirely based on the image modality, significantly different from other machine translation tasks.
\citet{tian-etal-2023-image} construct a dataset containing images of one-line texts with black font and white background, improving the translation quality of IIMT with a segmented pixel transformer.
\citet{lan2024translatotronv} adopt the ViT-VQGAN to IIMT with OCR and TIT multi-task learning, constructing a dataset with a single color as a background with multi-line texts.
\citet{qian-etal-2024-anytrans} utilize Qwen and AnyText models to translate images with short text on a real dataset but do not release their dataset publicly.

\paragraph{Text-within-Image Generation}
The research of text-within-image generation aims to generate clear and readable text in the images, which is generally necessary to incorporate textual priors, such as text position and image with rendered text \cite{ma2023glyphdraw, tuo2024anytext, zhang2023brush}.
\citet{Rodriguez_2023_WACV} propose OCR-VQGAN, which uses OCR pre-trained features to calculate the text perceptual loss, to mitigate the phenomenon of the generated blurring text with VQGAN \cite{Esser_2021_CVPR}.
And it is generally necessary to incorporate textual priors into the model, such as text position and image with rendered texts \cite{ma2023glyphdraw, tuo2024anytext, zhang2023brush}.

\section{Data Construction}
\label{sec:data}
In previous works of IIMT, \citet{mansimov-etal-2020-towards, tian-etal-2023-image} construct images with black font and white background of one-line texts. \citet{lan2024translatotronv} render multi-line texts into single-color backgrounds. 
The backgrounds of the above datasets are simple, and there is a gap between the constructed datasets and reality.
Therefore, we build a dataset containing real-world images as backgrounds, which is lacking in IIMT research. The images in our dataset are more complex, which are shown in Figure \ref{fig:example1}(b).

We build the dataset with images and texts from Multi30k\footnote{\url{https://github.com/multi30k/dataset}} \cite{W16-3210}. The Multi30k dataset is designed for Multimodal Machine Translation, and there are captions in several languages for each image.
Firstly, we resize the images in Multi30k to $512 \times 512$. 
Secondly, the German and English text captions are rendered
into the image by Pillow\footnote{\url{https://pypi.org/project/pillow/}} library.
Finally, each image is cropped to a size of $48 \times 512$, thereby fully preserving the text area and part of the background.

Our constructed IIMT30k dataset comprises three subsets, each characterized by a specific font: Times New Roman (TNR), Arial, and Calibri.
The texts in each subset of images are exclusively presented in one of these three fonts,
namely IIMT30k-TNR, IIMT30k-Arial, and IIMT30k-Calibri. Each subset is further divided into training, valid, and test sets.
The purpose of using multiple fonts is to investigate the font adaption capability of the model. 
We perform inspections on our dataset, ensuring the integrity of texts in the images.
Although the dataset is synthetic, it is comparable with real-world video subtitle images, and samples from the dataset are shown in Appendix \ref{appendix:samples}.

\section{Method}
\label{sec:model} 
\begin{figure}[h]
    \centering
    \includegraphics[width=\linewidth]{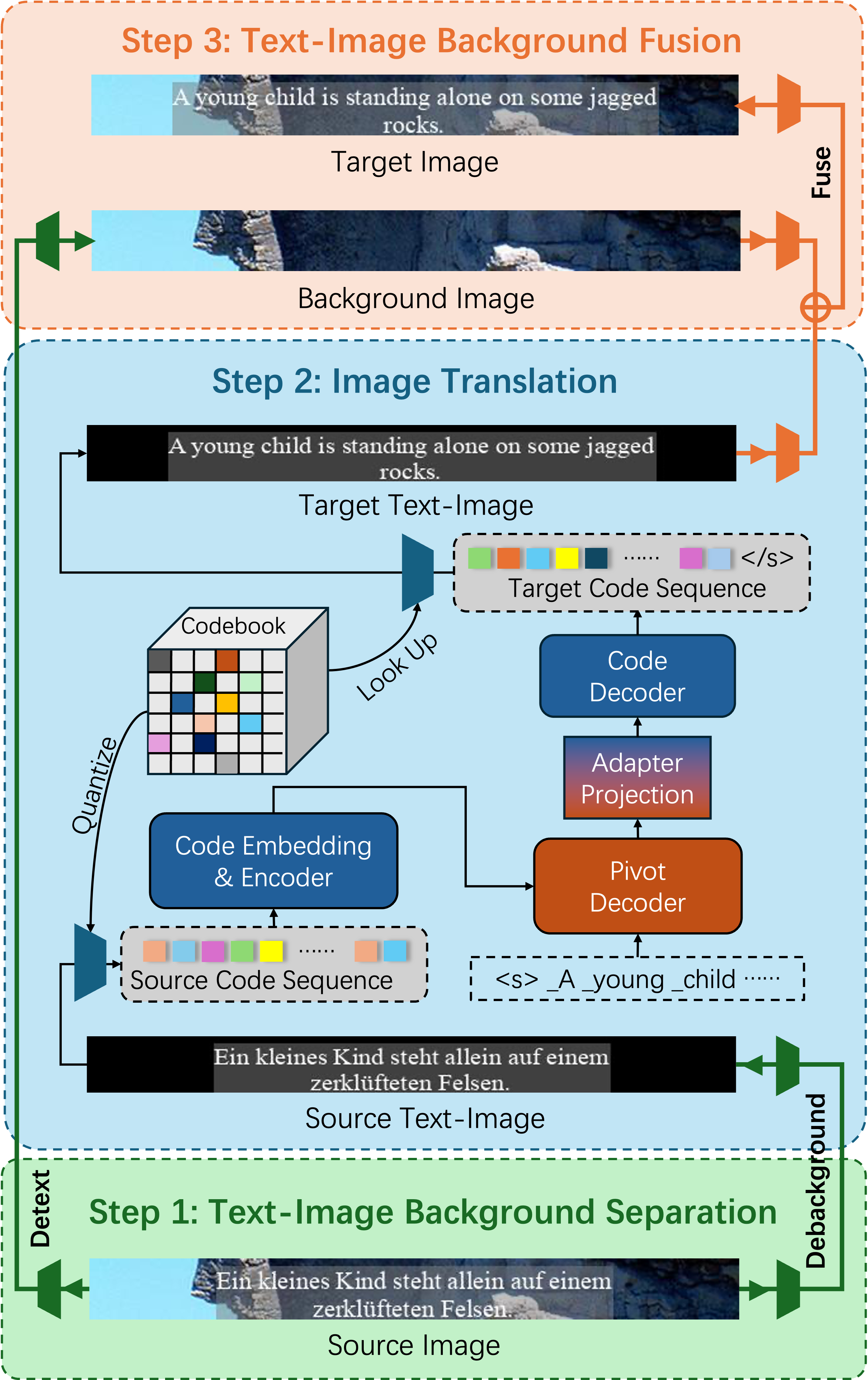}
    \caption{Architecture of our proposed DebackX.}
    \label{fig:model}
\end{figure}

We design the DebackX model which is shown in Figure \ref{fig:model}, and our model contains three components, the Text-Image Background Separation, the Image Translation, and the Text-Image Background Fusion.
The Text-Image Background Separation model first decomposes the source image into a background image and a source text-image. Then the Image Translation model transforms the source text-image into the target text-image. Finally, the Text-Image Background Fusion model fuses the background image and target text-image, generating the target image.

\subsection{Text-Image Background Separation}
The Text-Image Background Separation model decomposes the source image into a background image and a source text-image. 
The input source image $x$ is encoded with two ViT \cite{dosovitskiy2021image} encoders, $E_\text{deback}$ and $E_\text{detext}$, obtaining corresponding features separately. 
Then two ViT decoders $G_\text{back}$ and $G_\text{text}$ take the features as inputs, outputting the source text-image and the background image. Formally as:
\begin{align*}
    \text{Background Image}&: G_\text{back}(E_\text{deback}(x)), \\
    \text{Source Text-Image}&: G_\text{text}(E_\text{detext}(x)).    
\end{align*}

\subsection{Image Translation}
The Image Translation model is used to transform source text-images into target text-images.
The source text-image $x$ is encoded to feature by a ViT encoder $E$, and tokenized with a codebook $q$ into $z=(z_1, z_2, \ldots, z_N)$. 
The codebook $q$ contains $V$ learnable vectors $\{e_1, e_2, \ldots, e_V\}$, and each encoded feature $x_i$ is quantized by the nearest vector in $q$, obtaining $z_i$. Formally as:
\begin{equation}
    z_i = q(E(x_i)) = \argmin_{e_k \in q} || E(x_i) - e_k ||_2.
\end{equation}

Therefore, the images can be represented using the indices of the vectors in the codebook.
By tokenizing the source and target text-images into code sequences, the continuous representation of the image is transformed into a discrete sequence. Consequently, the image-to-image translation task shifts to a transformation from source code sequence to target code sequence.
Then, the source code sequence is passed through a Code Embedding layer and a Code Encoder in turn, generating the representation $H^E_{\text{code}}$.

To incorporate the semantic information required for the translation task into the code sequence, a Pivot Decoder is trained with TIT auxiliary task.
The target language text shifted to the right and prefixed with a Beginning of Sentence (BOS) token, obtaining word embedding $E_{\text{TIT}}$ through a Text Embedding layer. 
The word embedding $E_{\text{TIT}}$ and the Code Encoder representation $H^E_{\text{code}}$ are jointly used as inputs to the Pivot Decoder, formally as ${\rm PivotDecoder}(E_{\text{TIT}}, H^E_{\text{code}}, H^E_{\text{code}})$\footnote{The input of the Transformer Decoder can generally be represented as ${\rm Decoder}(Q, K, V)$.}.

The output of the Pivot Decoder $H^D_{\text{pivot}}$ serves two purposes. 
First, it is projected to the vocabulary size through a linear layer for training the auxiliary TIT task. 
Second, it is used as input for the subsequent Code Decoder that generates the target code sequence.

For the Code Decoder, the target code sequence is first prefixed with a BOS token and then passed through the Code Embedding layer to obtain the vector representation $E_{\text{code}}$. The representation from the Pivot Decoder $H^D_{\text{pivot}}$ is passed through a Linear Adapter to produce $H^A_{\text{pivot}}$. 
Both $E_{\text{code}}$ and $H^A_{\text{pivot}}$ are used as inputs for the Code Decoder, formally as ${\rm CodeDecoder}(E_{\text{code}}, H^A_{\text{pivot}}, H^A_{\text{pivot}})$.

The output code sequence of the Code Decoder is transformed into vector representations by looking up in the codebook and then decoded into the target text-image using a ViT Decoder.

During the inference stage, the target text is decoded autoregressively with the source code sequence.
After the target text is completely decoded, the complete representation from the Pivot Decoder $H^D_{\text{pivot}}$ can be obtained, which is used for decoding the target code sequence autoregressively.

\subsection{Text-Image Background Fusion}
The Text-Image Background Fusion model fuses the background image and the target text-image which are generated by the Text-Image background Separation model and Image Translation model to the target image.
Two ViT encoders $E_\text{back}$ and $E_\text{text}$ are used to encode the background image $i_b$ and the target text-image $i_t$, obtaining the corresponding features.
The sum of the features is passed to a ViT decoder $G_\text{fuse}$ to get the final output of the model, formally as:

\begin{equation}
    \text{Target Image}: G_\text{fuse}(E_\text{back}(i_b) + E_\text{text}(i_t)).
\end{equation}

\section{Training}
\subsection{Text-Image Background Separation}
We adopt the training method of image generation task to train the Text-Image Background Separation model.
The reconstruction loss and perceptual loss\footnote{\url{https://github.com/richzhang/PerceptualSimilarity}} \cite{zhang2018unreasonable} are two commonly used criteria for image generation, and we use the combination of them as described in Equation \ref{eq:recloss}.
\begin{equation}
\label{eq:recloss}
    \mathcal{L}_\text{img}(y, \hat{y}) = ||y-\hat{y}||^2 +\lambda_\text{p}\mathcal{L}_\text{Perceptual}(y, \hat{y}),
\end{equation}
where $y$ is the generation result of the model and $\hat{y}$ is the ground truth image. $\lambda_\text{p}$ is loss weight, and it is set to $0.1$ in our experiments.

Both the background image $i_b$ and text-image $i_t$ generation use the loss function. 
The total loss function at this stage is the sum of those mentioned above, formally as $\mathcal{L}_\text{sep}=\mathcal{L}_\text{img}(i_b, \hat{i_b})+\mathcal{L}_\text{img}(i_t, \hat{i_t})$.

\subsection{Image Translation}
The training of our Image Translation model consists of two stages. 

\paragraph{Stage 1: Vector Quantization.} 
The codebook and ViT Encoder-Decoder which are used to tokenize and detokenize the images are trained with image reconstruction tasks.
We use the commitment loss for the training of the codebook, and the total loss function for Stage 1 is described as Equation \ref{eq:vqcloss}. 
Besides, Exponential Moving Average (EMA) is also used for updating the codebook, which is a more robust training method \cite{lancucki2020robust}.

\begin{equation}
\label{eq:vqcloss}
\begin{split}
    \mathcal{L}_\text{VQ}(y, \hat{y}) = ||y-\hat{y}||^2 + \lambda_\text{p}\mathcal{L}_\text{Perceptual}(y, \hat{y}) \\+ ||\text{sg}[z_q] - E(x)||_2^2,
\end{split}
\end{equation}
where $y$ is the generated image of the model and $\hat{y}$ is the ground truth image.
$||\text{sg}[z_q] - E(x)||_2^2$ is the commitment loss with stop-gradient operation $\text{sg[·]}$, and $z_q$ is the vector obtained by quantization using the codebook, while $x$ is the source image and $E(x)$ is the output feature of the ViT encoder which also serves as the input to the quantization layer.
$\lambda_\text{p}$ is loss weight, and it is set to $0.1$ in our experiments.

\paragraph{Stage 2: Translation.}
\label{sec:translation-train}
We train the translation model to transform the codes of source text-images to the codes of target text-images with the TIT auxiliary task. Both codes and auxiliary texts utilize cross-entropy with a label-smoothing factor of $0.1$ as the loss function, denoted as $\mathcal{L}_{\text{code}}$ and $\mathcal{L}_{\text{TIT}}$.
The total loss function of this stage is the sum of two components, formally as $\mathcal{L}_{\text{trans}} = \mathcal{L}_{\text{code}} + \mathcal{L}_{\text{TIT}}$.

In this stage, the Embedding Layers, the Code Encoder, the Pivot Decoder, the Code Decoder, the Adapter Projection, and Output Projections are trained together.
Besides, it should be noted that we construct additional text-images only for training at this stage.
The additional text-images are used to pre-train the model, while the IIMT30k training set is for fine-tuning. 

\subsection{Text-Image Background Fusion}
The Text-Image Background Fusion model takes the separated background image and the translated text-image as input, the target image as output, which is trained with the image generation loss function described as Equation \ref{eq:recloss}.

\begin{table*}[t]
    \centering
    \begin{tabular}{cccccccccc}
    \Xhline{1.5pt}
        \multirow{3}{*}{\textbf{Systems}} & \multicolumn{4}{c}{\textbf{Translation Quality (BLEU $\uparrow$ / COMET $\uparrow$)}} & \multicolumn{4}{c}{\textbf{Visual Effect (FID $\downarrow$)}} \\
        \cline{2-9}
        & \multicolumn{2}{c}{\textbf{De-En}} & \multicolumn{2}{c}{\textbf{En-De}} & \multicolumn{2}{c}{\textbf{De-En}} & \multicolumn{2}{c}{\textbf{En-De}} \\
        & \textbf{Valid} & \textbf{Test} & \textbf{Valid} & \textbf{Test} & \textbf{Valid} & \textbf{Test} & \textbf{Valid} & \textbf{Test} \\
        \hline
        VQGAN & 0.7 / 24.2 & 0.6 / 24.4 & 0.6 / 17.9 & 0.8 / 20.2 & 25.5 & 21.3 & 27.2 & 20.7 \\
        VAR & 0.1 / 22.5 & 0.1 / 22.2 & 0.3 / 18.3 & 0.2 / 17.7 & 21.2 & 15.7 & 20.1 & 14.1 \\
        MaskGIT & 0.2 / 23.7 & 0.2 / 24.0 & 0.1 / 18.9 & 0.2 / 18.5 & 21.5 & 20.9 & 22.2 & 18.8 \\
        TIT-Render & 13.8 / 53.1 & 12.1 / 49.6 & 14.0 / \textbf{46.3} & 10.2 / \textbf{43.9} & 137.4 & 133.2 & 121.7 & 119.0  \\
        PEIT-Render & 10.5 / 50.4 & 8.6 / 47.0 & 12.3 / 41.1 & 7.9 / 35.6 & 149.9 & 140.1 & 125.3 & 119.7  \\
        McTIT-Render & 14.2 / \textbf{53.5} & 11.7 / 47.3 & 13.5 / 43.3 & 10.5 / 42.8 & 141.2 & 137.5 & 124.9 & 117.5 \\
        Translatotron-V & 2.7 / 30.1 & 1.6 / 24.8 & 2.1 / 26.5 & 1.9 / 24.6 & 22.4 & 10.1 & 23.2 & 17.5 \\
        DebackX (ours) & \textbf{14.9} / 51.2 & \textbf{12.8} / \textbf{50.0} & \textbf{14.6} / 42.2 & \textbf{11.1} / 40.0 & \textbf{20.4} & \textbf{9.0} & \textbf{19.5} & \textbf{8.7} \\
    \Xhline{1.5pt}
    \end{tabular}
    \caption{Experimental results of different systems. Metrics include translation quality (BLEU, COMET) and visual effect (FID). $\uparrow$ or $\downarrow$ indicates higher or lower values are better.}
    \label{tab:mainresult}
\end{table*}

\section{Experiments}

\subsection{Metrics}
It is hard to evaluate the translation quality of IIMT directly, so we first recognized the texts in the generated images with the OCR model and calculated the BLEU \cite{papineni2002bleu} and COMET \cite{rei-etal-2020-comet} scores between the OCR results and the reference texts, which is a commonly used method.
We use EasyOCR\footnote{\url{https://github.com/JaidedAI/EasyOCR}}, a widely used OCR toolkit that supports both German and English to recognize texts in the output images. 
BLEU and COMET scores are calaulted by Sacrebleu\footnote{\url{https://github.com/mjpost/sacrebleu}} and Unbabel-COMET\footnote{\url{https://github.com/Unbabel/COMET}} respectively.

To evaluate the visual effect of the output images, we use a widely used metric Fréchet Inception Distance (FID) \cite{heusel2018gans}, which is a measure of similarity between two images, calculated by computing the Fréchet distance between two Gaussians fitted to feature representations of the Inception network.
FID is shown to correlate well with the human judgment of visual quality and it is most often used to evaluate the quality of generated images \cite{lucic2018gans}. In our experiments, we employ pytorch-fid\footnote{\url{https://github.com/mseitzer/pytorch-fid}} to calculate FID scores. 

\subsection{Experimental Settings}
From the perspective of input and output modalities, IIMT is an image-to-image generation task. 
Therefore, we adopt several image-to-image generation models \cite{Esser_2021_CVPR, tian2024visual, chang2022maskgit} for the IIMT task.
Besides, we build models consisting TIT models \cite{zhu-etal-2023-peit, lan2023exploring} and Rendering.
We also investigate the performance of the latest IIMT model Translatotron-V \cite{lan2024translatotronv}.
The experiments are conducted on the IIMT30k-TNR dataset. The training set includes $25,205$ image pairs containing German and English parallel texts, and the valid set includes $864$ image pairs, while the test set includes $2,740$ image pairs. The following is a brief introduction to each system.

\paragraph{VQGAN~\cite{Esser_2021_CVPR}.} A system that tokenizes images using a codebook and generates images using an autoregressive Transformer Decoder. In our experiments, we train the codebook using all the German and English images and train the Transformer Decoder models separately for each translation direction.

\paragraph{VAR~\cite{tian2024visual}.} An improved system based on VQGAN, utilizing a more efficient and effective autoregressive generation paradigm namely ``next-resolution prediction''.

\paragraph{MaskGIT~\cite{chang2022maskgit}.} An image-to-image generation system that uses a bidirectional transformer decoder for synthesizing high-fidelity and high-resolution images. It predicts masked tokens in all directions during training and refines images iteratively during inference.

\paragraph{TIT-Render.} A pipeline system contains TIT model and text rendering. In our experiments, we use Transformer based image encoder and text decoder to complete the TIT task, obtaining the translated texts in the source images.

\paragraph{PEIT-Render.} An advanced TIT model PEIT \cite{zhu-etal-2023-peit} and text rendering pipeline.

\paragraph{McTIT-Render.} An OCR-NMT TIT model McTIT \cite{lan2023exploring} and text rendering pipeline.

\paragraph{Translatotron-V~\cite{lan2024translatotronv}.} An IIMT model adopts the architecture of ViT-VQGAN, with OCR and TIT multi-task training.

\paragraph{DebackX.} Our IIMT model introduced in Section \ref{sec:model}. The implementation details are introduced in Appendix \ref{appendix:implement} and Appendix \ref{appendix:training}.

\subsection{Main Results}
The experimental results of different systems are shown in Table~\ref{tab:mainresult}.
The autoregressive-based image-to-image generation model (VQGAN) can produce relatively clear characters while maintaining a sharp background. However, it fails to generate complete words and semantically meaningful phrases in the image, resulting in poor translation quality. 
The image-to-image generation models that do not adopt conventional autoregressive (VAR and MaskGIT) produce relatively clear backgrounds but fail to generate clear characters, resulting in even worse translation quality.

The methods consisting of TIT models and Rendering (TIT-Render, PEIT-Render, and McTIT-Render) can obtain meaningful translation results, but the rendering procedure removes the areas containing texts, significantly affecting the visual effect.
The Translatotron-V is designed for IIMT with a simple background, but cannot adapt to IIMT with complex scenarios, resulting in poor translation quality.

Compared with other IIMT methods, our system specifically designs an Image Translation model, achieving a better translation quality. Furthermore, the Text-Image Background Separation and Fusion models lead to the improvement of visual effects.

\section{Analysis}
IIMT produces visual output, requiring evaluation of both translation quality and visual effect. In this section, we conduct the following analytical experiments focusing on the above evaluation aspects:

\paragraph{Translation quality.} (1) We analyze the effectiveness of the pre-training followed by the fine-tuning strategy adopted in the Image Translation model and demonstrate how pre-training contributes to improvements in translation quality. (2) We conduct an ablation study to investigate the contribution of each component in our model to the translation quality.

\paragraph{Visual effect.} (3) Maintaining font consistency is crucial for the visual effect of translated outputs, yet this aspect is not captured by the used FID metric directly. To address the lack of font adaptation in current IIMT research, we use a dataset containing multiple fonts to evaluate whether our DebackX model can handle this challenge. (4) We conduct a case study to illustrate how the output of our model differs from other IIMT models, including TIT-Render and GPT-4o. Visually demonstrate the advantages of our DebackX model.

\subsection{Pre-training Study}
\label{sec:pre-training}
Due to the size of the dataset which consists of images and their captions in multiple languages is small, it is hard to build a large amount of images with complex backgrounds to train IIMT models directly. 
However, text-images used to train the Image Translation model are easy to construct with parallel corpus. Other IIMT systems cannot utilize such data, which is an advantage of our model. 
We construct $100$K German-English text-image pairs with the IWSLT dataset used to pre-train the Image Translation model (``+IWSLT PT'').
It should be noted that we use this system in the main experiments (Table \ref{tab:mainresult}) since its parameter is comparable to other systems.
Moreover, we construct $1$M text-image pairs with WMT14 datasets for pre-training (``+WMT14 PT''). To match the parameter of the model with the $1$M size of the dataset, we use a larger size of the model. 

\begin{table}[h]
    \centering
    \begin{tabular}{ccccc}
    \Xhline{1.5pt}
    \multirow{2}{*}{\textbf{Training Sets}} & \multicolumn{2}{c}{\textbf{De-En}} & \multicolumn{2}{c}{\textbf{En-De}} \\
    & \textbf{Valid} & \textbf{Test} & \textbf{Valid} & \textbf{Test} \\
    \hline
    IIMT30k-TNR & 7.4 & 5.9 & 8.1 & 5.4 \\
    +IWSLT PT & 14.9 & 12.8 & 14.6 & 11.1 \\ 
    +WMT14 PT & 20.6 & 17.5 & 18.9 & 14.4 \\
    \Xhline{1.5pt}
    \end{tabular}
    \caption{BLEU scores on different training sets, ``IIMT30k-TNR'' refers to only training with the IIMT30k-TNR training set, while ``+IWSLT PT'' and ``+WMT14 PT'' refer to pre-training with the corresponding dataset firstly and then fine-tuning on IIMT30k-TNR training set.}
    \label{tab:pre-training}
\end{table}

The experimental results are shown in Table \ref{tab:pre-training}, and results demonstrate the effectiveness of pre-training for the Image Translation Model. By increasing the size of the dataset, the translation quality is improved.

\subsection{Ablation Study}
We conduct an ablation study to further investigate the contribution of each part in our model.
Firstly, we remove the Pivot Decoder, which also means there is no TIT auxiliary task (\#2), and the Image Translation model learns to transform the code sequence directly. 
Secondly, we remove both the Text-Image Background Separation and Fusion models (\#3), training a new image-to-image translation model that takes the source image as input and produces the target image as output directly, which is similar to the two-pass model \cite{inaguma-etal-2023-unity}, and there is no procedure of debackground in the model. 
Finally, we remove both parts above mentioned (\#4).

Since the experiment \#3 and \#4 cannot utilize the large number of text-images for pre-training, to ensure the comparability of the experimental results, we do not use any pre-training data in the ablation study. 

\begin{table}[h]
    \centering
    \begin{tabular}{ccccccc}
    \Xhline{1.5pt}
    \multirow{2}{*}{\small{\textbf{\#}}} & \multirow{2}{*}{\small{\textbf{Pivot}}} & \multirow{2}{*}{\small{\textbf{Deback}}} & \multicolumn{2}{c}{\textbf{De-En}} & \multicolumn{2}{c}{\textbf{En-De}} \\
    &&& \small{\textbf{Valid}} & \small{\textbf{Test}} & \small{\textbf{Valid}} & \small{\textbf{Test}} \\
    \hline
    1 & \ding{51} & \ding{51} & 7.4 & 5.9 & 8.1 & 5.4 \\
    2 & \ding{55} & \ding{51} & 1.5 & 1.4 & 1.6 & 1.4 \\
    3 & \ding{51} & \ding{55} & 1.2 & 1.1 & 1.3 & 1.3 \\
    4 & \ding{55} & \ding{55} & 0.6 & 0.6 & 0.7 & 0.9 \\
    \Xhline{1.5pt}
    \end{tabular}
    \caption{BLEU scores of ablation study, ``Pivot'' refers to the TIT auxiliary task with Pivot Decoder, and ``Deback'' refers to the Text-Image Background Separation and Fusion models.}
    \label{tab:ablation}
\end{table}

The experimental results are shown in Table \ref{tab:ablation}, which demonstrate that the Pivot Decoder, the Text-Image Background Separation, and Fusion models contribute to the translation quality.
Moreover, the results also indicate that using only a two-pass model leads to lower translation quality in complex scenarios IIMT.

\begin{table*}[t]
    \centering
    \begin{tabular}{cccc}
         \Xhline{1.5pt}
         \textbf{\#} & \textbf{Source Image (Input)} & \textbf{Systems} & \textbf{Target Image (Output)} \\
         \hline
         \multirow{4}{*}[-2ex]{1} & \multirow{4}{*}[-2ex]{\includegraphics[width=0.75\columnwidth]{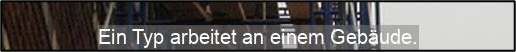}} & \raisebox{-.5\height}{Ground Truth} & \raisebox{-.5\height}{\includegraphics[width=0.75\columnwidth]{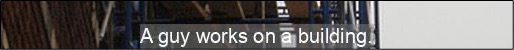}} \\
         \cline{3-4}
         & & \raisebox{-.5\height}{DebackX (Ours)} & \raisebox{-.5\height}{\includegraphics[width=0.75\columnwidth]{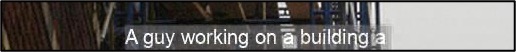}} \\
         \cline{3-4}
         & & \raisebox{-.5\height}{TIT-Render} & \raisebox{-.5\height}{\includegraphics[width=0.75\columnwidth]{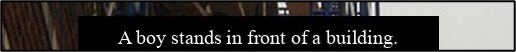}} \\
         \cline{3-4}
         & & \raisebox{-.5\height}{GPT-4o} & \raisebox{-.5\height}{\includegraphics[width=0.75\columnwidth]{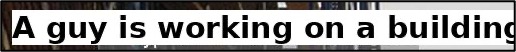}} \\
         \hline
         \multirow{4}{*}[-2ex]{2} & \multirow{4}{*}[-2ex]{\includegraphics[width=0.75\columnwidth]{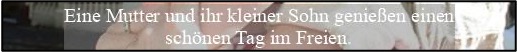}} & \raisebox{-.5\height}{Ground Truth} & \raisebox{-.5\height}{\includegraphics[width=0.75\columnwidth]{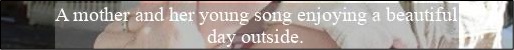}} \\
         \cline{3-4}
         & & \raisebox{-.5\height}{DebackX (Ours)} & \raisebox{-.5\height}{\includegraphics[width=0.75\columnwidth]{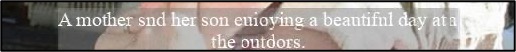}} \\
         \cline{3-4}
         & & \raisebox{-.5\height}{TIT-Render} & \raisebox{-.5\height}{\includegraphics[width=0.75\columnwidth]{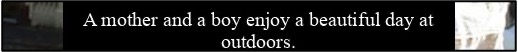}} \\
         \cline{3-4}
         & & \raisebox{-.5\height}{GPT-4o} & \raisebox{-.5\height}{\includegraphics[width=0.75\columnwidth]{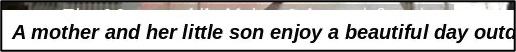}} \\
        \Xhline{1.5pt}
    \end{tabular}
    \caption{Comparison of the outputs of different systems, including our DebackX model, TIT-Render, and GPT-4o.}
    \label{tab:case}
\end{table*}

\subsection{Multiple Fonts Adaption Study} 
To investigate the adaption capability of multiple fonts, we mix the three subsets, IIMT30k-TNR, IIMT30k-Arial, and IIMT30k-Calibri. All the training sets, valid sets, and test sets consist of multiple font types.

The experimental results are shown in Table \ref{tab:fonts}, and the results illustrate that the datasets constructed with multiple fonts do not affect the translation quality significantly.
The reason why the BLEU scores for ``+Arial'' and ``+Arial Calibri'' are higher than that for ``TNR'' is the OCR model used for evaluation has a lower error rate in recognizing Arial and Calibri fonts in images compared to Times New Roman.

\begin{table}[h]
    \centering
    \begin{tabular}{ccccc}
    \Xhline{1.5pt}
    \multirow{2}{*}{\textbf{Training Sets}} & \multicolumn{2}{c}{\textbf{De-En}} & \multicolumn{2}{c}{\textbf{En-De}} \\
    & \textbf{Valid} & \textbf{Test} & \textbf{Valid} & \textbf{Test} \\
    \hline
    TNR & 7.4 & 5.9 & 8.1 & 5.4 \\
    +Arial & 10.4 & 8.5 & 9.2 & 6.5 \\
    +Arial Calibri & 10.8 & 8.6 & 9.5 & 6.9 \\
    \Xhline{1.5pt}
    \end{tabular}
    \caption{BLEU scores on training sets with multiple fonts. ``TNR'' refers to the dataset that is only constructed with Times New Roman. ``+Arial'' refers to the dataset that was constructed with 2 fonts, Times New Roman and Arial. ``+Arial Calibri'' refers to the dataset that was constructed with 3 fonts, Times New Roman, Arial, and Calibri.}
    \label{tab:fonts}
\end{table}

We conduct a manual font style recognition for the output images. 
If the font of characters in the output images differs from that in the input image, they are marked as inconsistent. 
After the manual evaluation, the font style consistency between the output images and the input images for ``+Arial'' is $97.3\%$, and for ``+Arial Calibri'' is $96.5\%$. 
Experimental results show that our model has the capability to adaption to multiple fonts.

\subsection{Case Study}
We illustrate the differences between the outputs of our DebackX model and TIT-Render with cases in Table \ref{tab:case}. Moreover, we conduct preliminary experiments with GPT-4o which is one of the best Multimodal Large Language Models, to test whether it can perform IIMT.

The input of Case \#1 contains German texts with Arial font and Case \#2 contains German texts with Times New Roman font.
For both Case \#1 and Case \#2, the output images of our DebackX can keep the same backgrounds as the source images, and the fonts of the text in the output images match the fonts in the input images.

The TIT-Render removes the text area in the source image, leading to the degradation of the visual effect severely. Since the Render procedure typically uses a fixed font for text rendering, it cannot ensure font consistency in Case \#1.

In our preliminary experiments, GPT-4o cannot complete the IIMT task directly. 
As a result, we decompose the IIMT task 
and design the following prompt for GPT-4o.
\textit{``Translate the German text into English,  erasing the text in this image and render the translated text into the processed image at the corresponding position.''}

GPT-4o generates the images based on its ``Analysis'', and the detailed outputs of GPT-4o are shown in Appendix \ref{appendix:appendix-gpt4o}.
For the outputs of GPT-4o, the issue with Case \#1 is that the translated text is not rendered in the corresponding position, occupying too much area.
The problem of Case \#2 is that the rendered text extends beyond the image boundary.
Moreover, the fonts in the output images are not consistent with the input images.
Although GPT-4o can generate meaningful translated text, it fails to comprehend the layout and font of the source image, resulting in poor visual quality in the output image.

\section{Conclusion}
In this paper, we propose the complex scenario IIMT task where the backgrounds are sourced from real-world images. 
To support the proposed task, we build a new IIMT dataset named IIMT30k, which comprises images with backgrounds derived from real-world. 
We introduce the DebackX model specifically designed to address the challenge of complex scenario IIMT.
Experimental results on the IIMT30k dataset demonstrate that our model outperforms previous IIMT models in both translation quality and visual effect.

\section*{Limitations}
While our DebackX model achieves a better performance on the IIMT task, this work has certain limitations.

Firstly, for the sub-modules of our model such as the vector quantization layer, the vision 
transformer encoder, and the decoder, we only conduct experiments using the most basic modules, lacking the investigation on other advanced modules.

Secondly, due to the multiple components in our model, which requires multi-stage training, leading to high computational resource costs.

\section*{Acknowledgment}
We thank all the anonymous reviewers for their insightful and valuable comments. This work is supported by the National Natural Science Foundation of China (Grant No. U21B2009, 62406015) and Beijing Institute of Technology Science and Technology Innovation Plan (Grant No. 23CX13027).

\bibliography{custom}

\begin{thebibliography}{28}
\providecommand{\natexlab}[1]{#1}

\bibitem[{Chang et~al.(2022)Chang, Zhang, Jiang, Liu, and Freeman}]{chang2022maskgit}
Huiwen Chang, Han Zhang, Lu~Jiang, Ce~Liu, and William~T. Freeman. 2022.
\newblock \href {https://arxiv.org/abs/2202.04200} {Maskgit: Masked generative image transformer}.
\newblock \emph{Preprint}, arXiv:2202.04200.

\bibitem[{Dosovitskiy et~al.(2021)Dosovitskiy, Beyer, Kolesnikov, Weissenborn, Zhai, Unterthiner, Dehghani, Minderer, Heigold, Gelly, Uszkoreit, and Houlsby}]{dosovitskiy2021image}
Alexey Dosovitskiy, Lucas Beyer, Alexander Kolesnikov, Dirk Weissenborn, Xiaohua Zhai, Thomas Unterthiner, Mostafa Dehghani, Matthias Minderer, Georg Heigold, Sylvain Gelly, Jakob Uszkoreit, and Neil Houlsby. 2021.
\newblock \href {https://arxiv.org/abs/2010.11929} {An image is worth 16x16 words: Transformers for image recognition at scale}.
\newblock \emph{Preprint}, arXiv:2010.11929.

\bibitem[{Elliott et~al.(2016)Elliott, Frank, Sima'an, and Specia}]{W16-3210}
Desmond Elliott, Stella Frank, Khalil Sima'an, and Lucia Specia. 2016.
\newblock \href {https://doi.org/10.18653/v1/W16-3210} {Multi30k: Multilingual english-german image descriptions}.
\newblock In \emph{Proceedings of the 5th Workshop on Vision and Language}, pages 70--74. Association for Computational Linguistics.

\bibitem[{Esser et~al.(2021)Esser, Rombach, and Ommer}]{Esser_2021_CVPR}
Patrick Esser, Robin Rombach, and Bjorn Ommer. 2021.
\newblock Taming transformers for high-resolution image synthesis.
\newblock In \emph{Proceedings of the IEEE/CVF Conference on Computer Vision and Pattern Recognition (CVPR)}, pages 12873--12883.

\bibitem[{Heusel et~al.(2018)Heusel, Ramsauer, Unterthiner, Nessler, and Hochreiter}]{heusel2018gans}
Martin Heusel, Hubert Ramsauer, Thomas Unterthiner, Bernhard Nessler, and Sepp Hochreiter. 2018.
\newblock \href {https://arxiv.org/abs/1706.08500} {Gans trained by a two time-scale update rule converge to a local nash equilibrium}.
\newblock \emph{Preprint}, arXiv:1706.08500.

\bibitem[{Inaguma et~al.(2023)Inaguma, Popuri, Kulikov, Chen, Wang, Chung, Tang, Lee, Watanabe, and Pino}]{inaguma-etal-2023-unity}
Hirofumi Inaguma, Sravya Popuri, Ilia Kulikov, Peng-Jen Chen, Changhan Wang, Yu-An Chung, Yun Tang, Ann Lee, Shinji Watanabe, and Juan Pino. 2023.
\newblock \href {https://doi.org/10.18653/v1/2023.acl-long.872} {{U}nit{Y}: Two-pass direct speech-to-speech translation with discrete units}.
\newblock In \emph{Proceedings of the 61st Annual Meeting of the Association for Computational Linguistics (Volume 1: Long Papers)}, pages 15655--15680, Toronto, Canada. Association for Computational Linguistics.

\bibitem[{Lan et~al.(2024)Lan, Niu, Meng, Zhou, Zhang, and Su}]{lan2024translatotronv}
Zhibin Lan, Liqiang Niu, Fandong Meng, Jie Zhou, Min Zhang, and Jinsong Su. 2024.
\newblock \href {https://arxiv.org/abs/2407.02894} {Translatotron-v(ison): An end-to-end model for in-image machine translation}.
\newblock \emph{Preprint}, arXiv:2407.02894.

\bibitem[{Lan et~al.(2023)Lan, Yu, Li, Zhang, Luan, Wang, Huang, and Su}]{lan2023exploring}
Zhibin Lan, Jiawei Yu, Xiang Li, Wen Zhang, Jian Luan, Bin Wang, Degen Huang, and Jinsong Su. 2023.
\newblock \href {https://arxiv.org/abs/2305.17415} {Exploring better text image translation with multimodal codebook}.
\newblock \emph{Preprint}, arXiv:2305.17415.

\bibitem[{Loshchilov and Hutter(2019)}]{loshchilov2019decoupled}
Ilya Loshchilov and Frank Hutter. 2019.
\newblock \href {https://arxiv.org/abs/1711.05101} {Decoupled weight decay regularization}.
\newblock \emph{Preprint}, arXiv:1711.05101.

\bibitem[{Lucic et~al.(2018)Lucic, Kurach, Michalski, Gelly, and Bousquet}]{lucic2018gans}
Mario Lucic, Karol Kurach, Marcin Michalski, Sylvain Gelly, and Olivier Bousquet. 2018.
\newblock \href {https://arxiv.org/abs/1711.10337} {Are gans created equal? a large-scale study}.
\newblock \emph{Preprint}, arXiv:1711.10337.

\bibitem[{Ma et~al.(2022)Ma, Zhang, Tu, Han, Wu, Zhao, and Zhou}]{ma2022improving}
Cong Ma, Yaping Zhang, Mei Tu, Xu~Han, Linghui Wu, Yang Zhao, and Yu~Zhou. 2022.
\newblock \href {https://arxiv.org/abs/2210.03887} {Improving end-to-end text image translation from the auxiliary text translation task}.
\newblock \emph{Preprint}, arXiv:2210.03887.

\bibitem[{Ma et~al.(2023{\natexlab{a}})Ma, Zhang, Tu, Zhao, Zhou, and Zong}]{ma-etal-2023-ccim}
Cong Ma, Yaping Zhang, Mei Tu, Yang Zhao, Yu~Zhou, and Chengqing Zong. 2023{\natexlab{a}}.
\newblock \href {https://doi.org/10.18653/v1/2023.findings-emnlp.330} {{CCIM}: Cross-modal cross-lingual interactive image translation}.
\newblock In \emph{Findings of the Association for Computational Linguistics: EMNLP 2023}, pages 4959--4965, Singapore. Association for Computational Linguistics.

\bibitem[{Ma et~al.(2023{\natexlab{b}})Ma, Zhao, Chen, Wang, Niu, Lu, and Lin}]{ma2023glyphdraw}
Jian Ma, Mingjun Zhao, Chen Chen, Ruichen Wang, Di~Niu, Haonan Lu, and Xiaodong Lin. 2023{\natexlab{b}}.
\newblock \href {https://arxiv.org/abs/2303.17870} {Glyphdraw: Seamlessly rendering text with intricate spatial structures in text-to-image generation}.
\newblock \emph{Preprint}, arXiv:2303.17870.

\bibitem[{Mansimov et~al.(2020)Mansimov, Stern, Chen, Firat, Uszkoreit, and Jain}]{mansimov-etal-2020-towards}
Elman Mansimov, Mitchell Stern, Mia Chen, Orhan Firat, Jakob Uszkoreit, and Puneet Jain. 2020.
\newblock \href {https://doi.org/10.18653/v1/2020.nlpbt-1.8} {Towards end-to-end in-image neural machine translation}.
\newblock In \emph{Proceedings of the First International Workshop on Natural Language Processing Beyond Text}, pages 70--74, Online. Association for Computational Linguistics.

\bibitem[{Papineni et~al.(2002)Papineni, Roukos, Ward, and Zhu}]{papineni2002bleu}
Kishore Papineni, Salim Roukos, Todd Ward, and Wei-Jing Zhu. 2002.
\newblock Bleu: a method for automatic evaluation of machine translation.
\newblock In \emph{Proceedings of the 40th annual meeting of the Association for Computational Linguistics}, pages 311--318.

\bibitem[{Qian et~al.(2024)Qian, Zhang, Yang, Fan, Ma, Wong, Sun, and Ji}]{qian-etal-2024-anytrans}
Zhipeng Qian, Pei Zhang, Baosong Yang, Kai Fan, Yiwei Ma, Derek~F. Wong, Xiaoshuai Sun, and Rongrong Ji. 2024.
\newblock \href {https://doi.org/10.18653/v1/2024.findings-emnlp.137} {{A}ny{T}rans: Translate {A}ny{T}ext in the image with large scale models}.
\newblock In \emph{Findings of the Association for Computational Linguistics: EMNLP 2024}, pages 2432--2444, Miami, Florida, USA. Association for Computational Linguistics.

\bibitem[{Rei et~al.(2020)Rei, Stewart, Farinha, and Lavie}]{rei-etal-2020-comet}
Ricardo Rei, Craig Stewart, Ana~C Farinha, and Alon Lavie. 2020.
\newblock \href {https://doi.org/10.18653/v1/2020.emnlp-main.213} {{COMET}: A neural framework for {MT} evaluation}.
\newblock In \emph{Proceedings of the 2020 Conference on Empirical Methods in Natural Language Processing (EMNLP)}, pages 2685--2702, Online. Association for Computational Linguistics.

\bibitem[{Rodr{\'\i}guez et~al.(2023)Rodr{\'\i}guez, Vazquez, Laradji, Pedersoli, and Rodriguez}]{Rodriguez_2023_WACV}
Juan~A. Rodr{\'\i}guez, David Vazquez, Issam Laradji, Marco Pedersoli, and Pau Rodriguez. 2023.
\newblock Ocr-vqgan: Taming text-within-image generation.
\newblock In \emph{Proceedings of the IEEE/CVF Winter Conference on Applications of Computer Vision (WACV)}, pages 3689--3698.

\bibitem[{Sennrich et~al.(2016)Sennrich, Haddow, and Birch}]{sennrich-etal-2016-neural}
Rico Sennrich, Barry Haddow, and Alexandra Birch. 2016.
\newblock \href {https://doi.org/10.18653/v1/P16-1162} {Neural machine translation of rare words with subword units}.
\newblock In \emph{Proceedings of the 54th Annual Meeting of the Association for Computational Linguistics (Volume 1: Long Papers)}, pages 1715--1725, Berlin, Germany. Association for Computational Linguistics.

\bibitem[{Tian et~al.(2024)Tian, Jiang, Yuan, Peng, and Wang}]{tian2024visual}
Keyu Tian, Yi~Jiang, Zehuan Yuan, Bingyue Peng, and Liwei Wang. 2024.
\newblock \href {https://arxiv.org/abs/2404.02905} {Visual autoregressive modeling: Scalable image generation via next-scale prediction}.
\newblock \emph{Preprint}, arXiv:2404.02905.

\bibitem[{Tian et~al.(2023)Tian, Li, Liu, Guo, and Wang}]{tian-etal-2023-image}
Yanzhi Tian, Xiang Li, Zeming Liu, Yuhang Guo, and Bin Wang. 2023.
\newblock \href {https://doi.org/10.18653/v1/2023.findings-emnlp.1004} {In-image neural machine translation with segmented pixel sequence-to-sequence model}.
\newblock In \emph{Findings of the Association for Computational Linguistics: EMNLP 2023}, pages 15046--15057, Singapore. Association for Computational Linguistics.

\bibitem[{Tuo et~al.(2024)Tuo, Xiang, He, Geng, and Xie}]{tuo2024anytext}
Yuxiang Tuo, Wangmeng Xiang, Jun-Yan He, Yifeng Geng, and Xuansong Xie. 2024.
\newblock \href {https://arxiv.org/abs/2311.03054} {Anytext: Multilingual visual text generation and editing}.
\newblock \emph{Preprint}, arXiv:2311.03054.

\bibitem[{Vaswani et~al.(2023)Vaswani, Shazeer, Parmar, Uszkoreit, Jones, Gomez, Kaiser, and Polosukhin}]{vaswani2023attention}
Ashish Vaswani, Noam Shazeer, Niki Parmar, Jakob Uszkoreit, Llion Jones, Aidan~N. Gomez, Lukasz Kaiser, and Illia Polosukhin. 2023.
\newblock \href {https://arxiv.org/abs/1706.03762} {Attention is all you need}.
\newblock \emph{Preprint}, arXiv:1706.03762.

\bibitem[{Yu et~al.(2022)Yu, Li, Koh, Zhang, Pang, Qin, Ku, Xu, Baldridge, and Wu}]{yu2022vectorquantized}
Jiahui Yu, Xin Li, Jing~Yu Koh, Han Zhang, Ruoming Pang, James Qin, Alexander Ku, Yuanzhong Xu, Jason Baldridge, and Yonghui Wu. 2022.
\newblock \href {https://openreview.net/forum?id=pfNyExj7z2} {Vector-quantized image modeling with improved {VQGAN}}.
\newblock In \emph{International Conference on Learning Representations}.

\bibitem[{Zhang et~al.(2023)Zhang, Chen, Wang, Lu, and Qiao}]{zhang2023brush}
Lingjun Zhang, Xinyuan Chen, Yaohui Wang, Yue Lu, and Yu~Qiao. 2023.
\newblock \href {https://arxiv.org/abs/2312.12232} {Brush your text: Synthesize any scene text on images via diffusion model}.
\newblock \emph{Preprint}, arXiv:2312.12232.

\bibitem[{Zhang et~al.(2018)Zhang, Isola, Efros, Shechtman, and Wang}]{zhang2018unreasonable}
Richard Zhang, Phillip Isola, Alexei~A. Efros, Eli Shechtman, and Oliver Wang. 2018.
\newblock \href {https://arxiv.org/abs/1801.03924} {The unreasonable effectiveness of deep features as a perceptual metric}.
\newblock \emph{Preprint}, arXiv:1801.03924.

\bibitem[{Zhu et~al.(2023)Zhu, Li, Lei, and Xiong}]{zhu-etal-2023-peit}
Shaolin Zhu, Shangjie Li, Yikun Lei, and Deyi Xiong. 2023.
\newblock \href {https://doi.org/10.18653/v1/2023.acl-long.751} {{PEIT}: Bridging the modality gap with pre-trained models for end-to-end image translation}.
\newblock In \emph{Proceedings of the 61st Annual Meeting of the Association for Computational Linguistics (Volume 1: Long Papers)}, pages 13433--13447, Toronto, Canada. Association for Computational Linguistics.

\bibitem[{Łańcucki et~al.(2020)Łańcucki, Chorowski, Sanchez, Marxer, Chen, Dolfing, Khurana, Alumäe, and Laurent}]{lancucki2020robust}
Adrian Łańcucki, Jan Chorowski, Guillaume Sanchez, Ricard Marxer, Nanxin Chen, Hans J. G.~A. Dolfing, Sameer Khurana, Tanel Alumäe, and Antoine Laurent. 2020.
\newblock \href {https://arxiv.org/abs/2005.08520} {Robust training of vector quantized bottleneck models}.
\newblock \emph{Preprint}, arXiv:2005.08520.

\end{thebibliography}

\appendix

\section{Details of DebackX}
\label{appendix:implement}
Our implementation for all ViT encoders and decoders is referring to timm\footnote{\url{https://github.com/huggingface/pytorch-image-models}}. 
We use learnable position encoding, and the patch embedding of the encoder uses a convolutional layer with a kernel size and stride equal to the patch size, while the output layer of the decoder uses a transposed convolutional layer with a kernel size and stride equal to the patch size.
All ViT encoders and decoders have patch\_size=$16$, d\_model=$512$, l=$8$, head=$8$, and d\_ff=$2,048$.

We implement the codebook based on vector-quantize-pytorch\footnote{\url{https://github.com/lucidrains/vector-quantize-pytorch}}, with a size of $8,192$ and a dimension of $32$. Therefore, two linear layers are placed both before and after the codebook to match the dimensions of the ViT features. This approach, known as Factorized Codes \cite{yu2022vectorquantized}, can improve the utilization of the codebook.

We use different model sizes for the Pivot Decoder and Code Decoder in the Image Translation model, 
since the auxiliary TIT task is easier than the task of code-to-code transformation. 
To prevent the auxiliary TIT task is overfitting while the other task is still underfitting, we use a smaller model size for Code Encoder and Pivot Decoder and a larger model size for Code Decoder. 
Both German and English texts for TIT auxiliary task are applied $10,000$ BPE \cite{sennrich-etal-2016-neural} to build a joint text vocabulary.

The detailed hyperparameters of experiments are shown in Table \ref{tab:param}.

\begin{table*}[t]
    \centering
    \begin{tabular}{cccccc|cccc}
    \Xhline{1.5pt}
    \multirow{2}{*}{\textbf{\#}} & \multirow{2}{*}{\textbf{Systems}} & \multicolumn{4}{c}{\textbf{Code Encoder \& Pivot Decoder}} & \multicolumn{4}{c}{\textbf{Code Decoder}} \\
    && \textbf{d\_model} & \textbf{l} & \textbf{head} & \textbf{d\_ff} & \textbf{d\_model} & \textbf{l} & \textbf{head} & \textbf{d\_ff} \\
    \hline
    1 & DebackX (Table \ref{tab:mainresult}) & 256 & 3 & 4 & 1,024 & 1,024 & 6 & 16 & 4,096 \\
    \hline
    2 & IIMT30k-TNR (Table \ref{tab:pre-training}) & 256 & 3 & 4 & 512 & 512 & 6 & 8 & 2,048 \\
    3 & +IWSLT PT (Table \ref{tab:pre-training}) & 256 & 3 & 4 & 1,024 & 1,024 & 6 & 16 & 4,096 \\
    4 & +WMT14 PT (Table \ref{tab:pre-training}) & 512 & 6 & 8 & 2,048 & 1,024 & 6 & 16 & 4,096 \\
    \hline
    5 & TNR (Table \ref{tab:fonts}) & 256 & 3 & 4 & 512 & 512 & 6 & 8 & 2,048 \\
    6 & +Arial (Table \ref{tab:fonts}) & 256 & 3 & 4 & 1,024 & 1,024 & 6 & 16 & 4,096 \\
    7 & +Arial Calibri (Table \ref{tab:fonts}) & 256 & 3 & 4 & 1,024 & 1,024 & 6 & 16 & 4,096 \\
    \Xhline{1.5pt}
    \end{tabular}
    \caption{Detailed hyperparameters for Image Translation model of our experiments.}
    \label{tab:param}
\end{table*}

\section{Training Details}
\label{appendix:training}
We use the Hugging Face Accelerate framework \footnote{\url{https://github.com/huggingface/accelerate}} with fp16 mixed precision to train our model. All parts of our models are trained with AdamW optimizer \cite{loshchilov2019decoupled}, inverse square root learning rate schedule, and dropout rate of $0.3$.

\paragraph{Text-Image Background Separation.} Two ViT encoders used to encode the source image and two ViT decoders used to decode the background image and text-image are trained for 25K steps.

\paragraph{Image Translation.} In \textbf{stage 1}, a ViT encoder is used to encode the text-image before the codebook, a ViT decoder is used to decode the text-image after the codebook, and the vectors containing in the codebook are trained for 50K steps together.
In \textbf{stage 2}, the additional text-images are used to pre-train the model for 100K steps, and the IIMT30k training set is used to fine-tune the model for 50K steps.

\paragraph{Text-Image Background Fusion.} Two ViT encoders are used to encode the background image and text-image, and a ViT decoder fusing two encoder features to decode the target image is trained for 15K steps.

\section{Samples from IIMT30k Dataset}
\label{appendix:samples}
The sizes of subsets in our dataset are shown in Table \ref{tab:data}.
The reason for the size difference in IIMT30k-Arial is that the Arial font occupies more space in the images, and we remove the images that do not fully render the text to ensure the integrity of the text in the images.

\begin{table}[h]
    \centering
    \begin{tabular}{cccc}
    \Xhline{1.5pt}
    Subsets & Training & Valid & Test \\
    \hline
    IIMT30k-TNR & 25,205 & 864 & 2,740 \\
    IIMT30k-Arial & 25,187 & 864 & 2,739 \\
    IIMT30k-Calibri & 25,205 & 864 & 2,740 \\
    \Xhline{1.5pt}
    \end{tabular}
    \caption{Statistic of IIMT30k dataset.}
    \label{tab:data}
\end{table}

\begin{figure*}[ht]
    \centering
    \includegraphics[width=\linewidth]{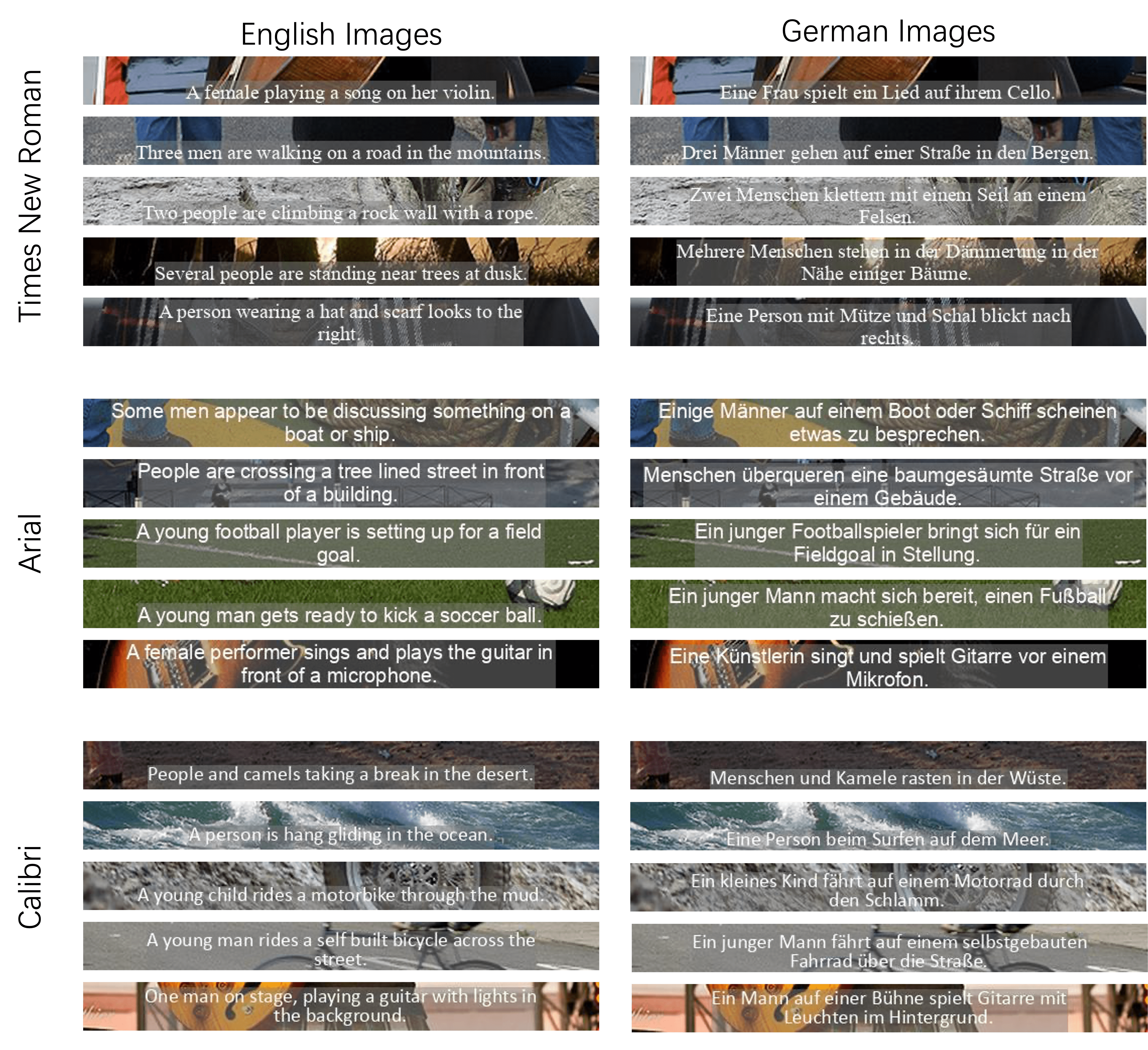}
    \caption{Samples from the IIMT30k dataset with multiple fonts (Times New Roman, Arial, and Calibri).}
    \label{fig:samples}
\end{figure*}

We sample some images with multiple fonts from the IIMT30k dataset, as shown in Figure \ref{fig:samples}.

\section{Impact of OCR Errors on Results Evaluation}
To investigate the impact of OCR errors on the evaluation of the results, we calculate the BLEU score and Word Error Rate (WER)\footnote{Calculate by jiwer library.} for the OCR recognized results of the golden output images with English (En) and German (De) texts. If there were no errors in the OCR recognized results, the BLEU score should be 100, and the WER should be 0. 

The results are shown in Table \ref{tab:ocrerrors}, and we test on both text-images and images with background (real images). The results show that even with the golden output images, the errors in OCR recognized results caused a certain decrease in the BLEU score.

\begin{table}[ht]
    \centering
    \begin{tabular}{cccccc}
    \Xhline{1.5pt}
    & & \multicolumn{2}{c}{\textbf{Text-Image}} & \multicolumn{2}{c}{\textbf{Real Image}} \\
    & & \textbf{En} & \textbf{De} & \textbf{En} & \textbf{De} \\
    \hline
    \multirow{2}{*}{BLEU} & Valid & 74.1 & 80.9 & 64.6 & 69.2 \\
    & Test & 76.0 & 81.0 & 67.5 & 69.2 \\
    \hline
    \multirow{2}{*}{WER} & Valid & 0.25 & 0.19 &0.32 & 0.30 \\
    & Test & 0.23 & 0.18 & 0.27 & 0.26 \\
    \Xhline{1.5pt}
    \end{tabular}
    \caption{BLEU score and WER of OCR recognized results for golden output images.}
    \label{tab:ocrerrors}
\end{table}

The results in Table \ref{tab:ocrerrors} also indicate that the OCR model has a lower error rate in recognizing text-images, suggesting that our Debackground method which separates the image into text-image and background is not only effective for IIMT but also has potential reference value for other tasks, such as OCR.

\section{Detailed Outputs of GPT-4o}
\label{appendix:appendix-gpt4o}
\begin{figure*}[ht]
    \centering
    \includegraphics[width=0.8\linewidth]{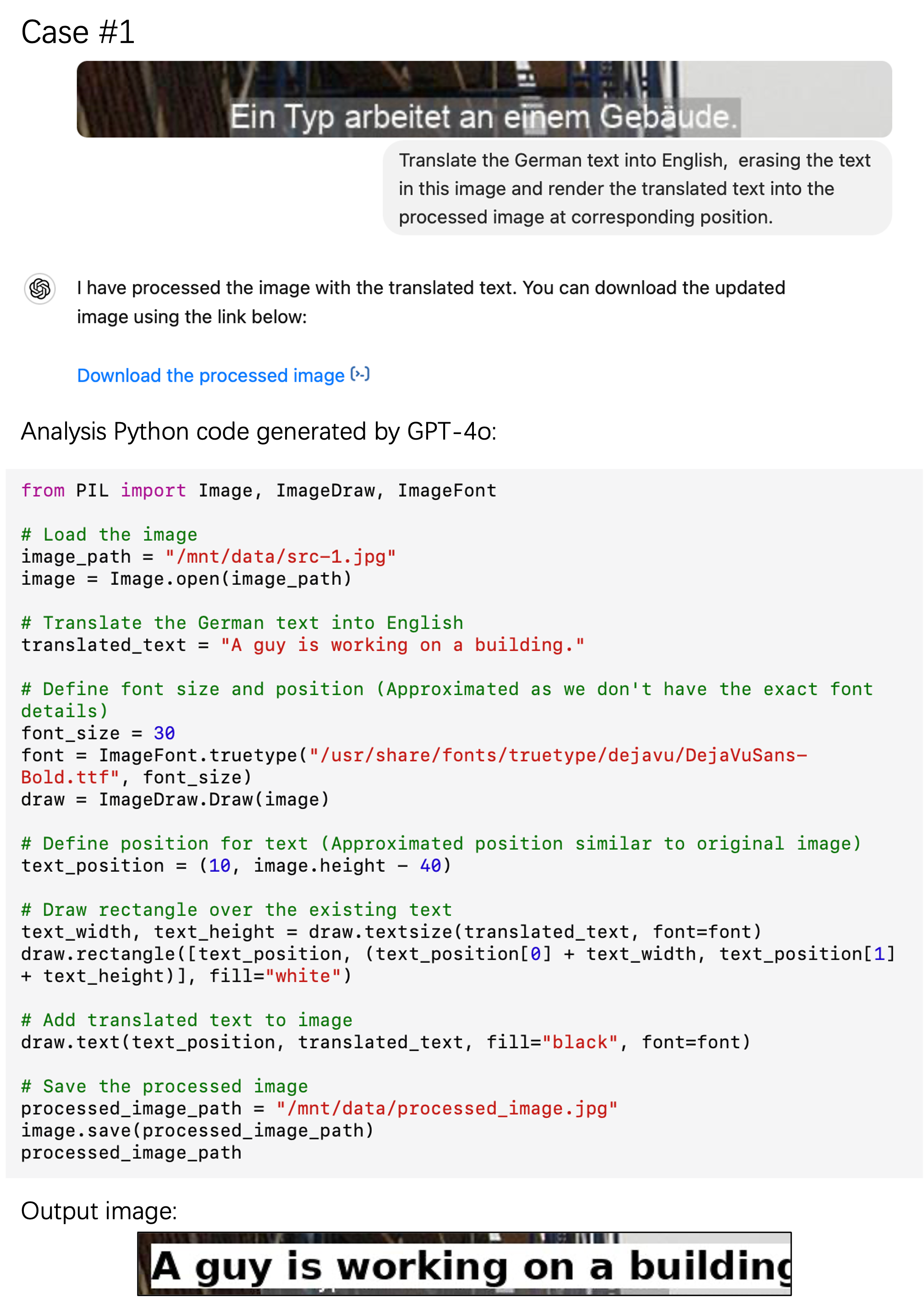}
    \caption{Detailed GPT-4o output of Case \#1.}
    \label{fig:gpt4odetail1}
\end{figure*}

\begin{figure*}[ht]
    \centering
    \includegraphics[width=0.8\linewidth]{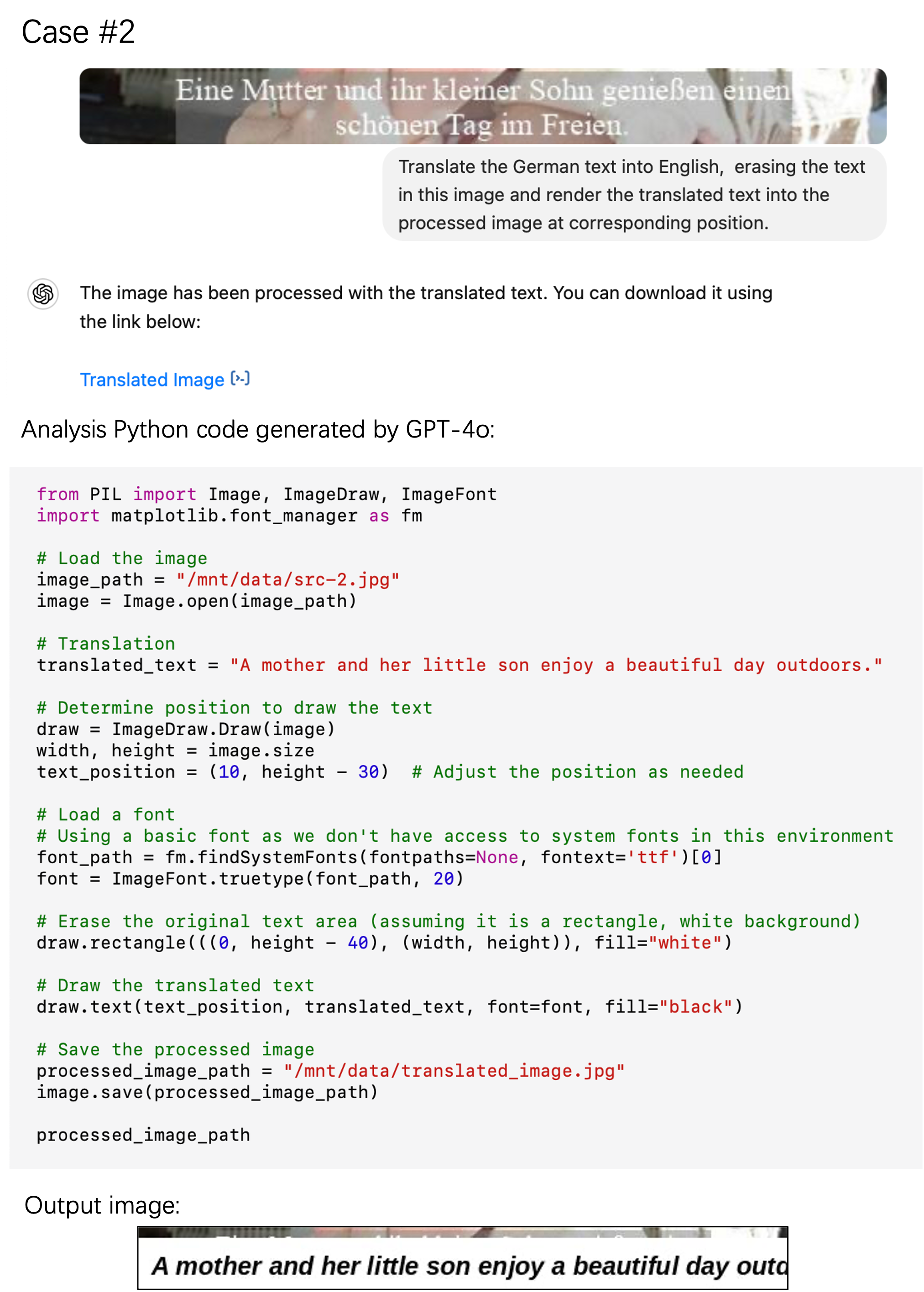}
    \caption{Detailed GPT-4o output of Case \#2.}
    \label{fig:gpt4odetail2}
\end{figure*}

The detailed outputs of GPT-4o that include Python code analysis of the case study are shown in Figure \ref{fig:gpt4odetail1} and Figure \ref{fig:gpt4odetail2}.
The prompt for GPT-4o is \textit{``Translate the German text into English, erasing the text in this image, and render the translated text into the processed image at the corresponding position.''}

\end{document}